\documentclass{article}

\PassOptionsToPackage{numbers}{natbib}
\usepackage[preprint]{iaseai26}

\usepackage[utf8]{inputenc} 
\usepackage[T1]{fontenc}    
\usepackage{hyperref}       
\usepackage{url}            
\usepackage{booktabs}
\usepackage{graphicx}
\usepackage{tabularx}
\usepackage[most]{tcolorbox}
\title{
  Mind the Gap! Pathways Towards Unifying AI Safety and Ethics Research
}
\author{%
  Dani Roytburg \\
  Department of Machine Learning \\
  Carnegie Mellon University\\
  Pittsburgh, PA 15213 \\
  \texttt{droytbur@andrew.cmu.edu} \\
  \And
  Beck Miller \\
  Department of Computer Science \\
  Emory University\\
  Atlanta, GA 30322 \\
  \texttt{bmill42@emory.edu} \\
}

\begin{document}

\maketitle

\begin{abstract}
  While much research in artificial intelligence (AI) has focused on scaling capabilities, the accelerating pace of development makes countervailing work on producing harmless, “aligned” systems increasingly urgent. Yet research on alignment has diverged along two largely parallel tracks: safety—centered on scaled intelligence, deceptive or scheming behaviors, and existential risk—and ethics—focused on present harms, the reproduction of social bias, and flaws in production pipelines. Although both communities warn of insufficient investment in alignment, they disagree on what alignment means or ought to mean. As a result, their efforts have evolved in relative isolation, shaped by distinct methodologies, institutional homes, and disciplinary genealogies. This fragmentation has become increasingly visible in both academic and public debates, even as the need to integrate technical and normative perspectives grows with each new milestone in AI scaling. We present the first \textbf{large-scale, quantitative evidence} of this schism through a bibliometric and network analysis of \textbf{6,442 papers} across twelve major machine learning and natural language processing conferences from 2020 to 2025. The results reveal a deeply \textbf{insular structure}: over \textbf{80\% of collaborations} occur within either safety or ethics, and researchers across the two communities are farther apart and statistically less reachable in the global co-authorship graph. Cross-disciplinary work is not only rare but structurally fragile—just \textbf{5\% of papers} are responsible for more than \textbf{85\% of all bridging connections}. Removing even a small number of these authors or papers dramatically increases network segregation, showing that cross-field collaboration depends on a handful of critical brokers rather than broad, systemic integration. These findings demonstrate that the safety–ethics divide is not merely rhetorical but \textbf{structural}, reflecting entrenched institutional silos that persist despite significant thematic overlap. The implications extend beyond academia: policy frameworks, governance models, and safety benchmarks increasingly mirror this same bifurcation, fragmenting what should be a unified effort toward human-compatible AI. For the IASEAI community—explicitly positioned at the intersection of safety, ethics, and alignment—our results underscore a defining challenge and opportunity. Achieving true alignment requires bridging the technical guarantees sought by safety research with the normative commitments advanced by ethics. Only through this synthesis can the field move beyond parallel concern toward a coherent discipline capable of producing systems that are not just powerful, but \textbf{responsible, robust, just, and safe}.\footnote{All code and filtered datasets will be made available upon acceptance.} 
\end{abstract}

\begin{figure}
    \centering
    \includegraphics[width=0.9\linewidth]{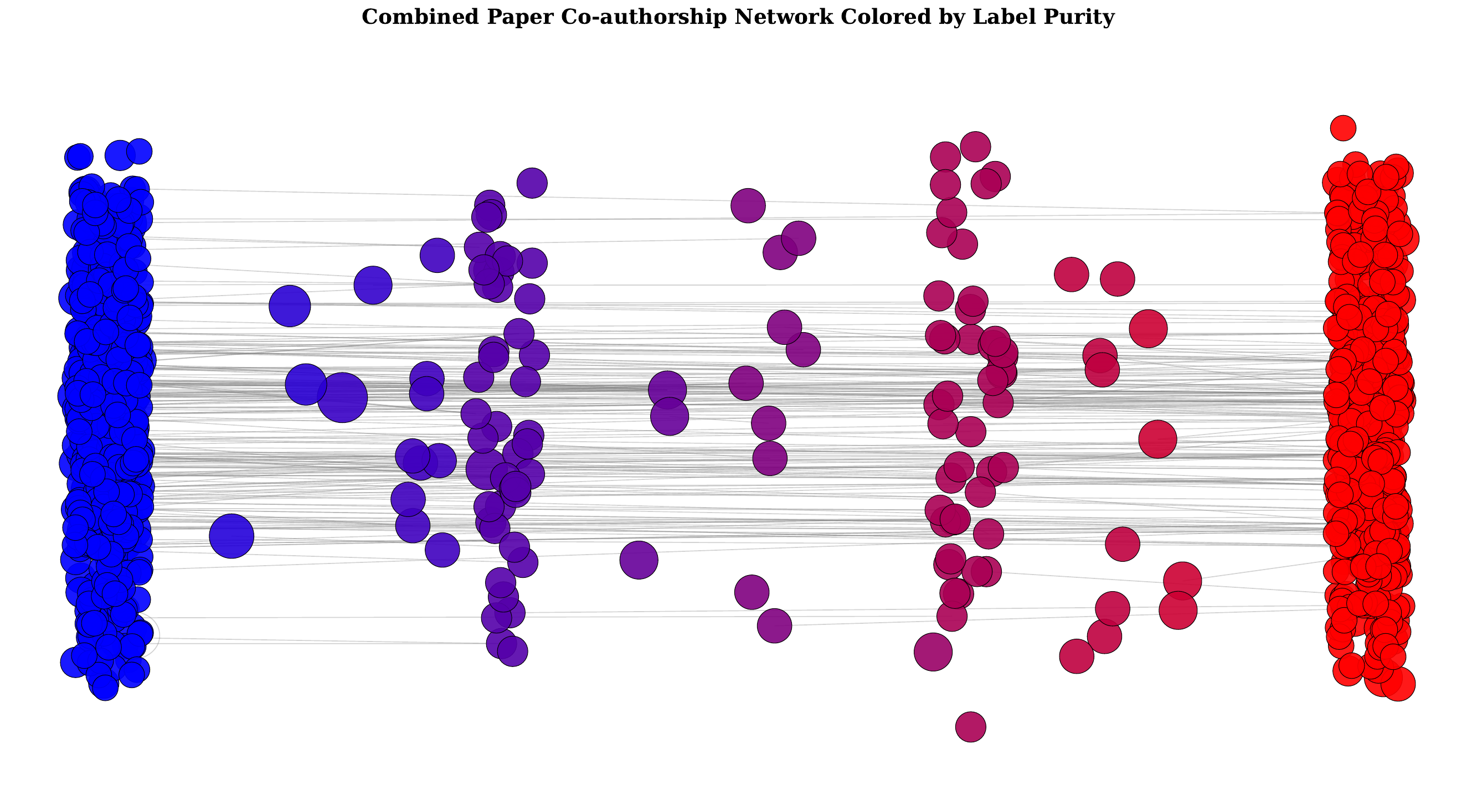}
    \caption{\textbf{Literature on AI Ethics} (left, in blue) and \textbf{AI Safety} (right, in red) is densely insular (83.1\% homophily), with a wide gap sparsely populated by \textbf{mixed-methods papers} (shades of purple). Sample of top 1000 nodes by degree.}
    \label{intro}
\end{figure}

\section{Introduction}
The need for \textit{both} safe and ethical artificial intelligence (AI) becomes more urgent with each new product release, research milestone, and high-stakes deployment. As corporations allocate increasing resources toward scaling capabilities—particularly through large language models (LLMs)—AI adoption appears poised to transform nearly every sector of modern economic and social life.

Most stakeholders would agree that an effective AI system should satisfy two desiderata: \textit{helpfulness}, the ability to achieve a specified goal, and \textit{harmlessness}, the ability to do so without generating negative consequences for individuals, organizations, or society.

What remains less clear is which desideratum should take precedence when helpfulness and harmlessness come into conflict. If a goal cannot be achieved without some risk of harm, should a system attempt completion or refusal? The range of contexts, risk tolerances, and normative perspectives that shape this dilemma illustrate what may be called a \textit{utility tradeoff} \cite{li2025smarterllmssaferexploring,shen2024jailbreak,vijjini-etal-2025-exploring}.

In practice, both theory and evidence suggest that free market forces bias heavily toward maximizing utility, since profitability depends more directly on capability than on restraint \cite{key62,elias_ai_2025}. Despite repeated open letters from leading scientists warning of the risks of unconstrained AI development, the largest technology firms continue to advance toward Artificial General Intelligence (AGI) with few structural checks on the prioritization of safety or ethics.

One might therefore expect researchers and advocates who emphasize harmlessness to collaborate extensively in counterbalancing this market-driven inertia. Yet, rather than converging on a unified strategy, these efforts reveal a growing \textit{schism}. Even among prominent voices, such as in a recent widely circulated TED Talk \cite{key19}, the divide between AI \textit{safety} and AI \textit{ethics} has become increasingly explicit.

This division is visible in the academic literature. The AI \textit{safety} community has historically emphasized technical guarantees around robustness, alignment, distributional shift, and long-term existential risk. The AI \textit{ethics} community, by contrast, has prioritized fairness, accountability, transparency, and the immediate social impacts of algorithmic systems. Both traditions share a concern with ensuring AI systems are aligned with human values, yet they diverge in methodologies, genealogies, and institutional homes. What results is a fragmented intellectual landscape where parallel efforts proceed with limited cross-pollination.

In this paper, we argue that greater integration of AI safety and AI ethics is both necessary and feasible. Our contributions are threefold:
\begin{enumerate}
    \item We systematize the distinct intellectual traditions of AI safety and ethics, identifying the core tensions that motivate our quantitative analysis.
    \item Using a network analysis of over 6,000 papers, we provide the first large-scale empirical evidence of the divide, measuring high homophily and fragile connectivity between the two fields.
    \item Drawing on our findings, we propose a concrete agenda for integrating the safety and ethics communities through shared research programs and venues.
\end{enumerate}

By combining conceptual analysis with network-based evidence, we provide a new foundation for understanding and addressing the structural divide between AI safety and AI ethics. For the IASEAI community—explicitly situated at the intersection of safety, ethics, and alignment—our results underscore both the urgency and the opportunity of building stronger bridges across these domains.

\section{Background}

\subsection{Framing}

We begin our analysis by grounding definitions of ``safety'' and ``ethics'' research in the intellectual traditions that have shaped them. Through a careful framing of these categories, we develop an intuition for their conceptual and methodological differences, yielding principles that inform our large-scale analysis of research silos. This allows us to interrogate what ``alignment'' means across distinct subfields that often share terminology but diverge in motivation and scope.

\subsection{A Sketch of AI Safety}

The AI Safety paradigm originates with philosophers like Nick Bostrom and Eliezer Yudkowsky, who analyze AGI through the lens of existential risk. Bostrom posits that the extreme polarity of potential outcomes—benevolent versus catastrophic ASI—necessitates preemptive alignment research \cite{bostrom_superintelligence_2014,key16}, whereas Yudkowsky argues that any superintelligence derived from current methods will be uncontrollable and lead to human extinction \cite{yudkowsky_if_2025}. Both converge on the urgency of ensuring robust human oversight.

These philosophical concerns were operationalized by \textit{Concrete Problems in AI Safety} \cite{amodei_concrete_2016}, which defined a technical research agenda around five problems: (1) Negative Side Effects, (2) Reward Hacking, (3) Scalable Oversight, (4) Safe Exploration, and (5) Robustness to Distributional Shift. From this agenda, key interventions emerged, including AI Alignment (outer/inner objective alignment \cite{key98}), AI Control (constraining harmful behaviors \cite{greenblatt2024aicontrolimprovingsafety}), and AI Interpretability (auditing opaque model mechanisms \cite{bereska2024mechanisticinterpretabilityaisafety}). The related field of AI Security adapts these concerns for present-day systems \cite{key41,shen2024jailbreak,lin_ai_2025,key73,key11}.

The field's trajectory is heavily influenced by Effective Altruism (EA), which frames existential risk mitigation as a long-term moral priority \cite{bostrom_superintelligence_2014}. This influence connects abstract risks to technical problems: ``negative side effects'' relates to Bostrom's paperclip problem, reward hacking to Yudkowsky's scenarios of deceptive goal-seeking \cite{yudkowsky_if_2025}, and ``scalable oversight'' to the challenge of supervising ASI \cite{key16}. Consequently, ``alignment'' is often treated as a formal property to be verified, using benchmarks and red-teaming as proxies for normative criteria, a practice critiqued as ``safetywashing'' \cite{ren2024safetywashingaisafetybenchmarks}.

\subsection{A Sketch of AI Ethics}

In contrast, AI Ethics is rooted in a socio-technical framework focused on immediate, systemic, and distributive harms \cite{borenstein_ai_2021}. Drawing from critical theory \cite{behrent2013foucault}, STS \cite{nath2023posthumanism}, and applied ethics, it prioritizes fairness, justice, and accountability in deployed systems \cite{key50}. These values are operationalized through frameworks like FATE (Fairness, Accountability, Transparency, and Ethics) \cite{key21}. \textbf{Fairness} research addresses embedded biases in data and models \cite{key24,key27,mehrabi2022surveybiasfairnessmachine,buolamwini2018gender,key102}; \textbf{Accountability} seeks legal and institutional loci of responsibility for AI-driven harms \cite{key28,novelli2024accountability,key66,judge_when_2025}; and \textbf{Transparency} aims to make automated decisions interpretable to affected stakeholders \cite{key24,key25,mosca2022shap}.

Where AI Safety often diagnoses failure as a mis-specified objective function, AI Ethics identifies it as the output of socio-technical systems that encode and scale historical inequities \cite{d2023data,buolamwini2018gender,benjamin2023race}. In this view, alignment is not about constraining an agent to a given objective, but about interrogating whose values and interests that objective represents. The field thus advocates for remedies beyond technical fixes, emphasizing participatory design, institutional governance, and robust regulatory oversight \cite{key66,judge_when_2025}.

\subsection{Convergences and Divergences}
\label{similar}

Despite divergent foundations, the communities share a foundational critique of unconstrained AI proliferation, evidenced by broad support for the 2023 research moratorium and the founding of labs like the \href{https://www.dair-institute.org/}{Distributed AI Research Institute} and \href{https://ssi.inc/}{Safe Superintelligence}. Promising frontiers for synthesis also exist in their technical agendas. For example, Transparency (Ethics) and Interpretability (Safety) both address model opacity, while Accountability (Ethics) and Scalable Oversight (Safety) both grapple with reliable human supervision of complex systems.

However, two fundamental tensions limit collaboration.

First, the \textbf{Distraction Argument} critiques the long-termist philosophy underpinning much of AI Safety \cite{bender2021dangers}. It posits that an overriding focus on low-probability, high-impact existential risks diverts finite resources—talent, funding, and attention—from immediate, realized harms disproportionately affecting marginalized communities \cite{swoboda_examining_2025,gebru_effective_nodate}. This creates a dynamic of ``deferred justice,'' where equity is postponed until speculative future risks are neutralized.

Second, the \textbf{Scoping Argument} raises concerns about the tractability of AI Ethics' recommendations. From an engineering perspective, its focus on systemic critique can produce recommendations that are difficult to formalize and implement within ML pipelines \cite{sadek2025challenges}. Calls for ``justice'' or ``inclusivity,'' while normatively crucial, may lack the technical specificity required for direct intervention in model architecture or training, posing a challenge for practitioners seeking actionable guidance \cite{munn2023uselessness}.

Rather than adjudicate these arguments, we posit they underlie the research silos discovered through our structural analysis, showcasing deep-seated epistemic tensions over risk, evidence, and moral priority. AI Safety frames alignment as a problem of control; AI Ethics frames it as a problem of justice. We conclude that these perspectives are not mutually exclusive but profoundly complementary. A truly safe system must be just, and a just system must be robust and controllable. Integrating these two research programs is a critical task for the future of artificial intelligence.

\begin{figure}
\centering
\includegraphics[width=0.75\linewidth]{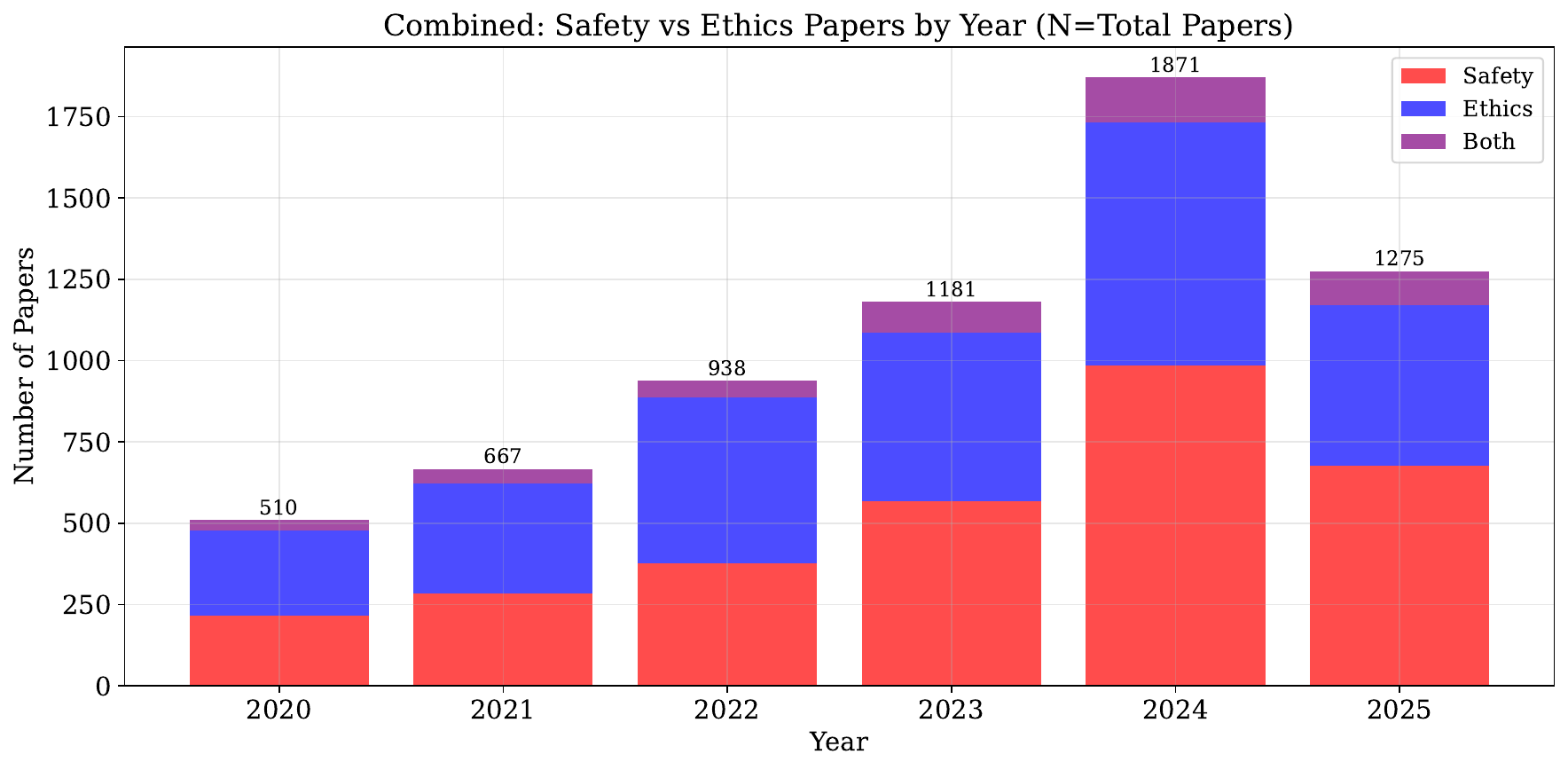}
\caption{Yearly growth of Ethics/Safety Research, 2020-2025}
\label{yearly}
\end{figure}
\section{Methods}
We seek to measure the impact that these tensions have on collaboration across the fields of AI Safety and Ethics research. Specifically, we analyze the structural interaction patterns of authors and their research at major machine learning and natural language processing conferences. Using graph-based network analysis, we conduct a set of tests which measure the connectivity between the two communities relative to their internal compositions. These tests are validated with graph-specific statistical significant tests such as rewiring and null-labeling.

\subsection{Data Collection, Filtering, and Preprocessing}
\subsubsection{Data Collection}
We collect the proceedings of all works published from 2020-2025 to 12 conference venues. These conferences were chosen due to their current centrality to research on artificial intelligence (4 conferences), natural language processing (5), or the particular domains of safety and ethics research in AI (3):

\textbf{Machine Learning Venues}: International Conference on Learning Representations (ICLR), International Conference on Machine Learning (ICML), Neural Information Processing Systems (NeurIPS), and the AAAI Conference on Artificial Intelligence (AAAI).

\textbf{Natural Language Processing Venues}: Meeting of the Association for Computational Linguistics (ACL), Nations of the Americas Chapter of the Association for Computational Linguistics (NAACL), Empirical Methods in Natural Language Processing (EMNLP), European Chapter of the Association for Computational Linguistics (EACL), Findings of ACL (Findings).

\textbf{Ethics and Safety-Specific Conferences}: AI Ethics and Society (AIES), Fairness, Accountability, and Transparency (FAccT), Secure and Trustworthy Machine Learning (SaTML).

We collect all papers submitted to these conferences, as well as archival workshops, demos, and tutorials which are published in the proceedings. Combined, this corpus constitutes 102,329 papers. For the natural language processing conferences, we collect the composite citations by filtering from the 2025 ACL Anthologies Dump \cite{acl2025dump}. For the machine learning and domain-specific conferences, we do the same using the DBLP 2025 XML dump \cite{dblp2025dump}. While ACL Anthologies contains all abstract information, the DBLP dump does not. Thus, we add abstracts for DBLP-derived entries by querying SemanticScholar, OpenAlex, and OpenReview (see Table \ref{tab:enrichment_coverage_yearly}). We do not include data from the 2025 NeurIPS and ICML proceedings, as they are not yet part of the official conference anthologies.

We limit our analysis to the scope of conferences, excluding preprints available on arXiv for several reasons. First, preprints do not go through the peer review process, and as such may not adequately represent community participation in the areas defined \cite{noauthor_analyzing_2019}. Second, arXiv submissions pose a double-counting risk as many authors submit their work to arXiv while awaiting submission feedback to conferences. Finally, since our study poses network-based hypotheses about broader research areas, searching all possible instances of representative work on arXiv would be infeasible without a network-based data generation process; this would distort our analysis since the basis of inclusion necessitates proximate network effects.

\subsubsection{Filtering}

After collecting our corpus of representative works, we proceed to filter out papers that do not relate to either AI Ethics or Safety research. First, we compose a set of 114 safety and 102 ethics keywords by hand, reading through several fundamental surveys in both communities to derive common terminology that would span the scope of representative research. We apply this filter to each conference. We pass papers which match at least one safety or ethics keyword to a second filter over where a language model decides if a paper matches our prompt, with a confidence score and reasoning. The keywords, LLM-filter configuration, and validation experiments are presented in Appendix \ref{filter_sec}. After filtering, we retain \textbf{6,442 papers} and \textbf{20,690 authors}.

\subsubsection{Preprocessing}

With our filtered dataset, we index authors and papers as belonging either to ethics, safety, or a mix. Specifically, we devise a score for each paper as the number of safety-specific keyword matches divided by the total number of ethics and safety keyword matches. Since each paper matches at least one keyword, this gives us ``pure'' papers, such that a ``pure'' AI Safety paper has a score of 1, while a pure Ethics paper has a score of 0. We denote these as ``pure'' papers, and classify anything in between with a ``mixed'' label.

For authors, we aggregate the scores of each of their papers and perform the same calculation. 

\begin{table}[htbp]
\centering
\caption{Classification of Papers and Authors in Filtered Dataset}
\label{tab:classification}
\begin{tabular}{@{}lrrrr@{}}
\toprule
Category & Pure Safety & Pure Ethics & Mixed & Total \\
\midrule
Papers & 2,874 (44.6\%) & 2,959 (45.9\%) & 609 (9.5\%) & 6,442 \\
Authors & 9,634 (46.5\%) & 7,264 (35.1\%) & 3,792 (18.3\%) & 20,690 \\
Authors ($\geq 2$ papers) & 1,233 (27.0\%) & 1,003 (22.0\%) & 2,328 (51.0\%) & 4,564 \\
\bottomrule
\end{tabular}
\end{table}

\subsection{Structural Network Analysis}

With this data, we now turn towards a co-authorship based analysis of community dynamics. Specifically, we analyze two types of networks: \textbf{Author Networks}, where each node denotes an author, and edges between authors are inversely weighted by the number of papers they co-authored; and \textbf{Paper Networks}, where each node denotes a paper and edges between papers are inversely weighted by the number of authors they have in common. For author-based networks, we exclude authors who only contribute to one paper, reducing our network to 4,564 authors (22.06\% of the filtered dataset). Without this filter, network effects become too sparse to analyze across aggregations, and our homophily and connectivity metrics would correlate with how many single-paper authors appear in ethics or safety-based networks.

Following best practices in bibliographic social network science, we study coauthorship dynamics instead of citation networks \cite{newman2004coauthorship}. We are that coauthorship networks demonstrate most directly the nature of collaboration and full engagement with the research agendas of co-authors. Additionally, the highly collaborative nature of AI research lends itself to dense networks of interactions, as demonstrated by the 3.2:1 author-to-paper ratio. Intuitively, coauthorship networks measure the first-order effects of propagating research through practice. In comparison, citation data comes with noise, imperfect signals of intellectual reception, and directionality---each of which are shown to pose unique challenges when trying to study the separation effects of research communities \cite{tahamtan2019citationcountsmeasureupdated}.

\subsubsection{Metrics}

With the author and paper-based networks, we conduct a suite of tests to assess structural separation between safety and ethics communities, following best practices \cite{newman2004finding}:

\paragraph{Homophily}: Fraction of edges (or edge weight) that connect authors of the same category (safety-safety, ethics-ethics) \cite{khanam2023homophily, horta_homophily_2022}. Weighted homophily accounts for co-authorship frequency. Here, we ask how many collaborations are exclusive to safety or ethics (author network), or how many papers are written by safety- or ethics-specific authors (paper network).
    
\paragraph{Bridge Concentration}: Fraction of Safety-Ethics shortest paths passing through a top-K set of authors (ranked by degree or centrality). Measures how concentrated cross-group brokerage is.
    
\paragraph{Average Shortest-Path (ASP)}: Median and mean shortest-path lengths for reachable pairs, both unweighted (hops) and weighted (distance = $\frac{1}{\texttt{weight}}$) . Computed for within-group (Safety-Safety, Ethics-Ethics) and cross-group (Safety-Ethics) pairs, with $\Delta = ASP_{S,E} - \texttt{mean}(ASP_{S,S}, ASP_{E,E})$ to measure relative separation. Weighted Average Shortest Path allows us to answer the question: for any two authors (or papers), how many papers-in-common (or authors-in-common) would it take to meet? This measures the distance of direct collaboration, providing a measure of the activity in research silos. We aim to demonstrate that aggregations of these ASPs show \textbf{more hops} and \textbf{greater distance} across disciplines.

\subsubsection{Statistical Significance Testing}

To ensure metrics reflect structural properties rather than artifacts, we employ multiple null models \cite{haghani_what_2023}:

\paragraph{Label-Shuffle Null}: Permute safety/ethics labels across fixed topology (500-2000 reps), recompute metrics, and compute empirical p-values and z-scores \cite{saxena2024shuffling}.
    
\paragraph{Degree-Preserving Rewire Null}: Randomize edges while preserving degree sequences, recompute metrics to test if topology alone explains separation \cite{vavsa2022null}.

These tests confirm that observed separation (e.g., high homophily, concentrated bridges, fragile reachability) is statistically significant and not due to random chance or degree distribution. Each null model was run with 1,000 permutations; p-values are estimated as the fraction of nulls exceeding observed values.

\section{Results and Discussion}
\subsection{Homophily}

Our analysis of \textbf{homophily}---the tendency for researchers to collaborate with others from their own community---provides the starkest evidence of the schism (See ``Baseline'' in Table \ref{homophily-table}). In our author network, we measured a global homophily of \textbf{83.1\%}. This indicates that \textbf{over four out of every five collaborations} occur between authors who are exclusively focused on either safety or ethics, demonstrating a powerful in-group preference. This insularity holds when examining each community individually: \textbf{73.5\%} of collaborations involving a safety researcher are with other safety specialists, while \textbf{68.2\%} of collaborations for ethics researchers are with their peers. These findings are statistically significant under shuffle-based and degree-preserving rewiring null models (p << 0.01 for both).

\begin{figure}[h!]
\centering
\includegraphics[width=0.9\linewidth]{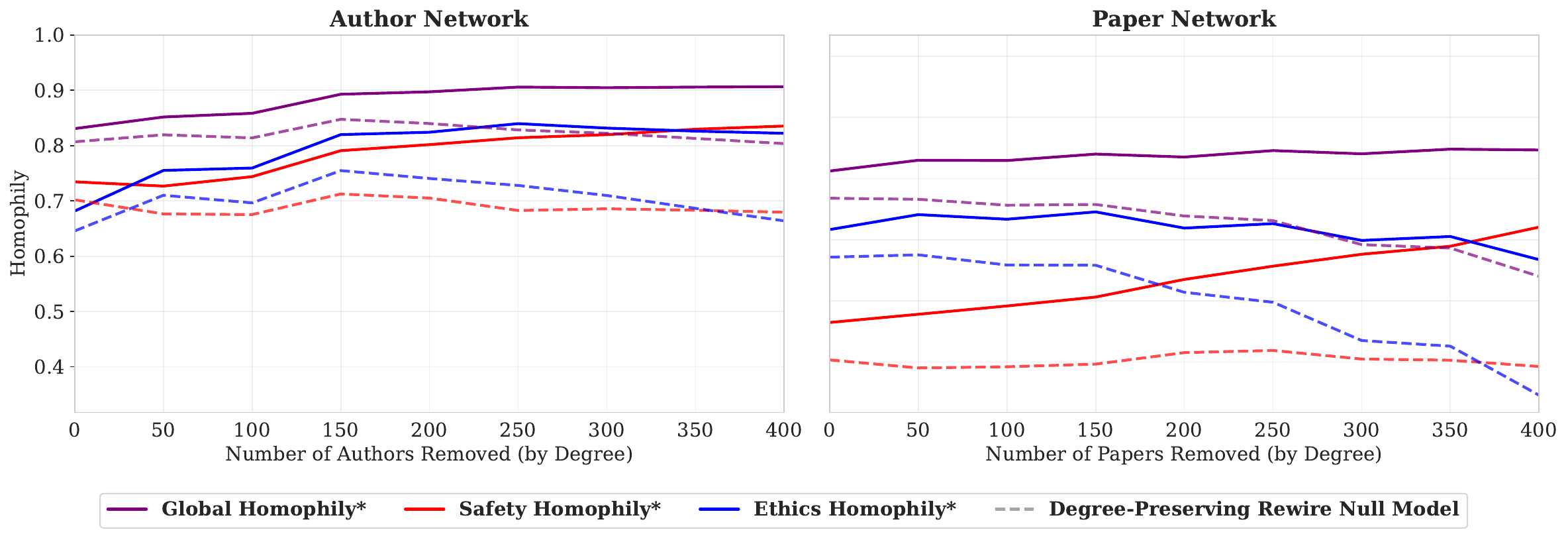}
\caption{Author-based (left) and paper-based (right) networks exhibit high homophily globally and in safety/ethics networks. They increase as top authors are pruned (p << 0.05).}
\label{homophily-chart}
\end{figure}

\begin{table}[ht]
\caption{Homophily values under degree-based removal of top authors}
\label{homophily-table}
\centering
\begin{tabular}{@{}l ccc ccc@{}}
\toprule
& \multicolumn{3}{c}{Author Network} & \multicolumn{3}{c}{Paper Network} \\
\cmidrule(r){2-4} \cmidrule(l){5-7}
\# Authors Removed & Global & Safety & Ethics & Global & Safety & Ethics \\
\midrule
\textbf{Baseline} (0) & 83.1\% & 73.5\% & 68.2\% & 71.2\% & 46.5\% & 61.6\% \\
100 & 85.8\% & 74.4\% & 75.9\% & 72.9\% & 49.2\% & 63.3\% \\
200 & 89.7\% & 80.2\% & 82.4\% & 73.5\% & 53.5\% & 61.9\% \\
300 & 90.5\% & 82.0\% & 83.2\% & 74.0\% & 57.6\% & 59.9\% \\
400 & 90.7\% & 83.5\% & 82.2\% & 74.7\% & 62.0\% & 56.7\% \\
\bottomrule
\end{tabular}
\end{table}

Interestingly, the effect is slightly less pronounced in our paper-based network (homophily of \textbf{71.2\%}), suggesting that while researchers themselves remain highly siloed, a small number of cross-disciplinary papers create connections that make the \textit{literature} appear more integrated than the \textit{social network} of its authors. This discrepancy highlights the outsized impact of a small number of \textbf{``bridge'' authors}. A single interdisciplinary author, by co-authoring one paper with the ``pure safety'' community and another with the ``pure ethics'' community, creates a direct link between those two otherwise disconnected bodies of literature. This single author's work acts as a hub, connecting entire clusters of papers and significantly lowering the paper network's homophily. The result reveals a critical insight: while the social network of researchers remains highly segregated, the literature itself is stitched together by the crucial, yet sparse, work of these bridging individuals.

To demonstrate this, we conduct a perturbation test by removing the nodes with the highest degrees and observing the effect on homophily (See  Figure \ref{homophily-chart}). In doing so, we show how the loss of well-connected bridge authors impacts total homophily. The author-based network shows a persistent increase, rising to 90.7\% global homophily. For papers, removing high-degree papers (papers written by prolific co-authors) contributes to a slight increase in homophily globally, but the effect is dispersed---while ethics research maintains its current rate with a slight decrease, the homophily of safety research \textit{increases} by 15\%. The divided impact suggests that the relative diversity of safety research is concentrated in ``hub'' papers---removing them quickly disconnects safety research from ethics-based or mixed-method literature.

\subsection{Bridge Connectivity}

The results in Table \ref{bridge-connectivity-table} reveal two critical structural properties of the safety-ethics network: \textbf{highly concentrated brokerage} and \textbf{extreme fragility}. At the baseline, we find that a small elite of authors---just the \textbf{top 1\%} by degree---act as brokers for a disproportionate \textbf{58.0\%} of all shortest paths between the two communities. This concentration is statistically significant (\(p<0.01\)) and indicates that cross-disciplinary communication relies on a "hub-and-spoke" model rather than a broad, distributed dialogue.

\begin{figure}[h!]
\centering
\includegraphics[width=0.8\linewidth]{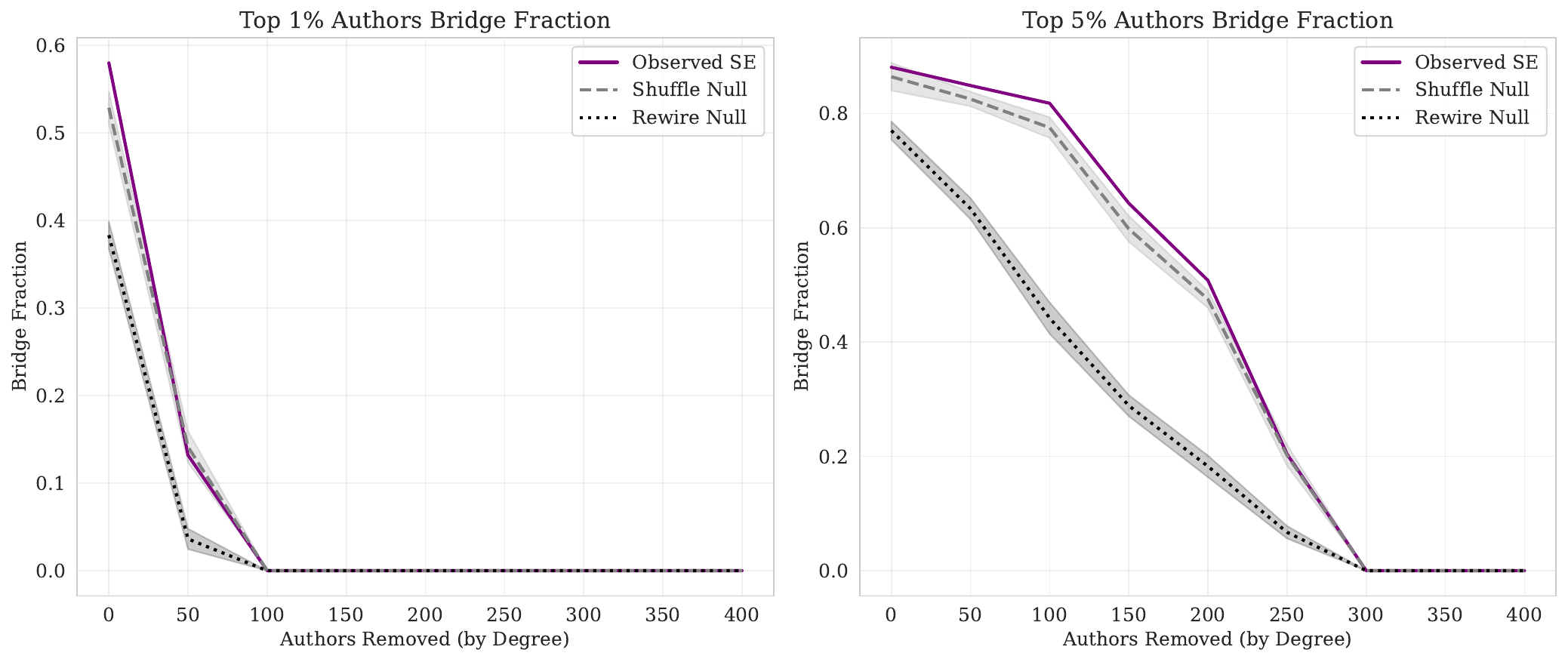}
\caption{Author-based bridge connectivity with authors removed by degree. 0 is the baseline, with shuffle and degree-preserving nulls included (p << 0.05).}
\label{bridge-connectivity-chart}
\end{figure}

Furthermore, this connectivity is exceedingly fragile. The targeted removal of these high-degree authors causes a precipitous drop in connectivity that far outpaces the decay in our null models. After removing just 100 authors, for instance, the most central bridge paths (those controlled by the top 1\%) vanish entirely (\textbf{0.0\%}), and overall connectivity plummets. This demonstrates that the intellectual exchange between AI safety and ethics is not a robust, resilient conversation but a tenuous one, critically dependent on a few central individuals.

\begin{table}[ht]
\caption{Fraction of bridge paths passing through top authors, under degree-based removal}
\label{bridge-connectivity-table}
\centering
\begin{tabular}{@{}l cc @{\hspace{1em}} cc @{\hspace{1em}} cc@{}}
\toprule
& \multicolumn{2}{c}{\textbf{Observed Network}} & \multicolumn{2}{c}{Label-Shuffle Null} & \multicolumn{2}{c}{Degree-Rewire Null} \\
\cmidrule(r){2-3} \cmidrule(lr){4-5} \cmidrule(l){6-7}
\# Authors Removed & Top 1\% & Top 5\% & Top 1\% & Top 5\% & Top 1\% & Top 5\% \\
\midrule
0 (Baseline) & \textbf{58.0\%} {\scriptsize(p<0.01)} & \textbf{88.1\%} {\scriptsize(p<0.05)} & 52.9\% & 86.4\% & 38.3\% & 77.0\% \\
50 & \textbf{13.2\%} {\scriptsize(p>0.05)} & \textbf{84.9\%} {\scriptsize(p<0.05)} & 14.1\% & 82.5\% & 3.6\% & 63.4\% \\
100 & \textbf{0.0\%} & \textbf{81.8\%} {\scriptsize(p<0.01)} & 0.0\% & 77.5\% & 0.0\% & 44.2\% \\
150 & \textbf{0.0\%} & \textbf{64.3\%} {\scriptsize(p<0.05)} & 0.0\% & 59.8\% & 0.0\% & 28.9\% \\
200 & \textbf{0.0\%} & \textbf{50.8\%} {\scriptsize(p<0.05)} & 0.0\% & 47.5\% & 0.0\% & 18.3\% \\
\bottomrule
\end{tabular}
\end{table}

\subsection{Weighted Average Shortest Path}

\begin{figure}[ht]

\centering

\includegraphics[width=0.8\textwidth]{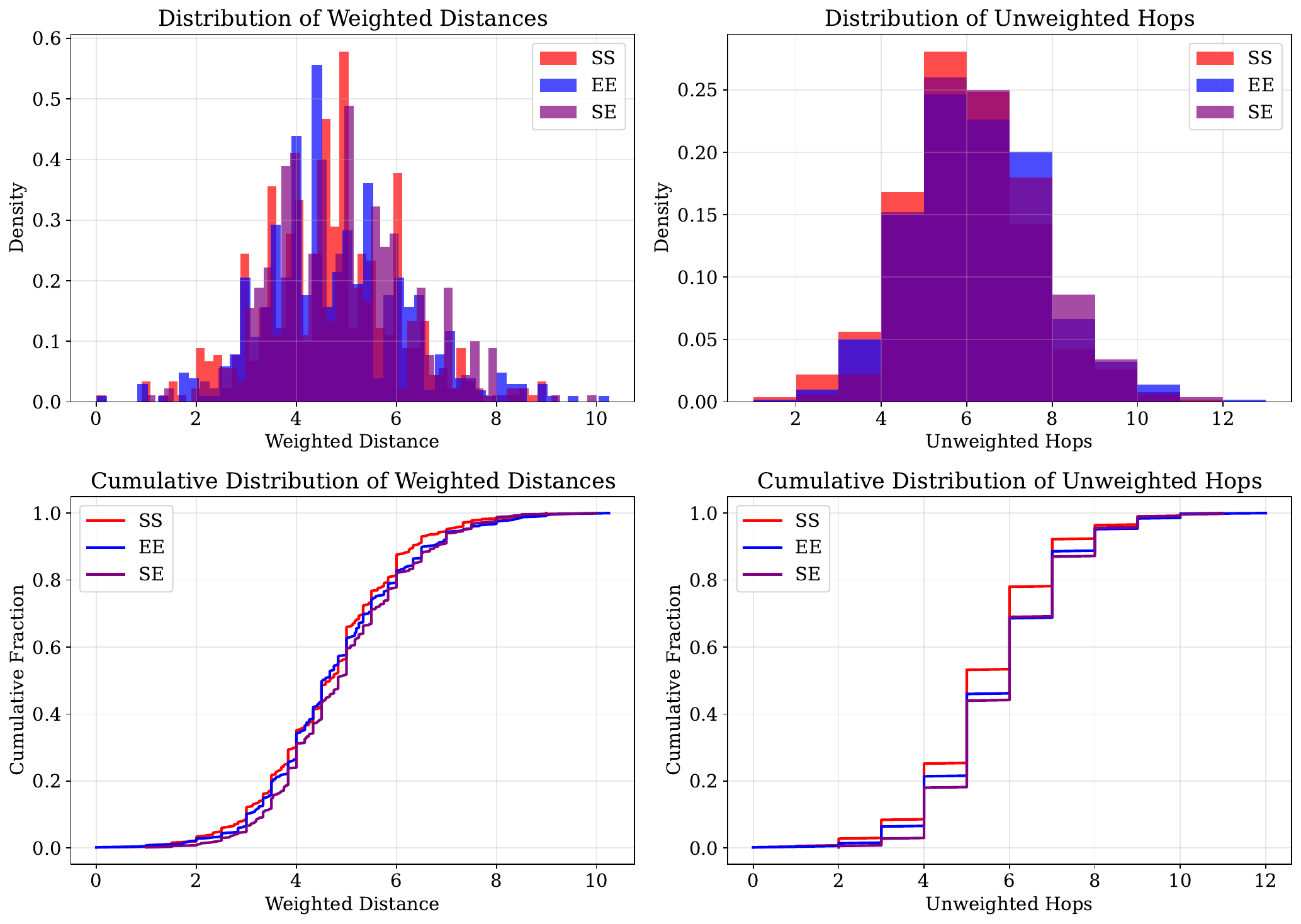}

\caption{Distributions of weighted path distances (left) and unweighted hops (right) for pairs of authors. The Safety-Ethics (SE) distribution is systematically shifted to the right compared to within-group Safety-Safety (SS) and Ethics-Ethics (EE) pairs, indicating longer paths are required to connect researchers across the two communities.}

\label{fig:wasp_distributions}

\end{figure}

Our analysis of the average shortest paths (ASP) between authors reveals a subtle but persistent "friction" that inhibits cross-disciplinary collaboration. While the mean path lengths are superficially similar, their underlying distributions, shown in Figure \ref{fig:wasp_distributions}, tell a clearer story. The cumulative distribution for cross-community (SE) paths is systematically shifted to the right, meaning a consistently longer weighted distance is required to connect a random pair of safety and ethics researchers compared to pairs within their own communities (\(p<0.01\)). For instance, \textbf{80\%} of all Safety-Safety author pairs can be connected within a weighted distance of 6.0, whereas reaching the same fraction of Safety-Ethics pairs requires a longer path.

This separation is further confirmed by our reachability analysis, which measures the fraction of author pairs connected within a given number of hops. We find that cross-group reachability is significantly lower than would be expected by chance. After five hops, only \textbf{16.9\%} of safety-ethics author pairs are connected, a figure substantially below the \textbf{21.5\%} we would expect in a randomly labeled network with the same structure (\(p<0.001\)). This demonstrates that the path to collaboration is not only longer on average but also structurally impeded, quantitatively confirming the schism between the two fields (see Table \ref{tab:wasp_reachability}).

\section{Limitations}
While this study provides the first large-scale, quantitative evidence of the schism between AI safety and ethics, we acknowledge several limitations that define the scope of our findings.

\subsection{Scoping and Data Representation}
Our analysis is scoped to the proceedings of 12 major academic conferences from 2020-2025. While these venues are central to the field, our corpus does not capture the full research ecosystem, which includes important work published in journals, workshops, industry technical reports, and influential preprints. Our focus on the post-2020 era captures the modern, LLM-centric landscape but excludes foundational work that shaped the communities prior to this period. Furthermore, while we provide a robust justification for excluding arXiv preprints, this choice means our analysis may not capture the most recent or fast-moving research trends that have yet to undergo peer review.

\subsection{Methodological Assumptions}
Our methodology relies on several key assumptions. First, our two-stage filtering process, while rigorously validated, necessarily operationalizes "safety" and "ethics" into discrete categories. These fields are complex and evolving, and a different conceptual taxonomy could yield different quantitative results. Second, we use co-authorship networks as a proxy for \textbf{active collaboration}. This is a strong signal of direct partnership but does not measure other forms of intellectual exchange, such as the passive influence captured by citation networks. Our findings are therefore a measure of social and institutional separation, not necessarily a complete lack of intellectual cross-pollination. Finally, our author network analysis focuses on researchers with at least two publications in our corpus. This standard technique allows for a robust analysis of the core research community but means our conclusions may be less representative of newcomers or authors with a single relevant publication.

\section{Conclusion}

The schism between AI safety and ethics is not merely anecdotal; it is a structural reality with measurable consequences. Our network analysis of over 6,000 papers provides quantitative evidence of this divide, revealing a landscape characterized by high homophily (\textbf{83.1\%} in-group collaboration) and fragile, concentrated brokerage. This structural gap persists despite significant thematic overlap, suggesting that the barriers to integration are more social and institutional than they are intellectual.

The consequences of this fragmentation extend beyond academia into global policy, creating a fractured approach to governance where our most pressing risks are treated as competing priorities. Landmark international reports on AI safety, for instance, have focused heavily on technical control and existential risk while largely neglecting the critical issues of bias and fairness central to the ethics community \cite{bengio_international_2025}. Conversely, influential frameworks like UNESCO's recommendations on AI ethics provide robust normative guidance on fairness but offer fewer tractable, technical solutions for implementation and do not engage with long-term catastrophic risks \cite{key32}. Each community, operating in isolation, exports an incomplete vision of "alignment" to policymakers.

It is precisely this gap that a venue like IASEAI is designed to fill. Its mission explicitly combines the language of safety, ethics, and alignment, creating a unique intellectual space for the cross-pollination our findings show is urgently needed. Yet, the very structure of our field's calls for papers---including IASEAI's---still tends to separate these topics into distinct tracks, underscoring how deeply ingrained this divide has become. Our work is a call to move beyond these inherited categories. We argue that the future of alignment research depends not on choosing between control and justice, but on synthesizing them into a unified, resilient, and truly effective discipline. We conclude with three pragmatic recommendations to accomplish such a future:

\paragraph{1. Shared Empirical Benchmarks.} The empirical standards of the two fields are almost entirely disjoint. Safety relies on adversarial red-teaming to probe for catastrophic failures, while ethics employs sociotechnical audits to uncover embedded biases \cite{ren2024safetywashingaisafetybenchmarks}. We advocate for the creation of \textbf{shared evaluative benchmarks} that treat alignment not as a bifurcated concept but as a unified property. For instance, a benchmark could require a model to demonstrate robustness to jailbreaking while simultaneously satisfying group fairness constraints, forcing a direct confrontation with the safety-utility tradeoffs that currently divide the fields.

\paragraph{2. Cross-Institutional Venues.} Our network analysis reveals a divide sustained by institutional geography—separate conferences, funding streams, and academic departments. To bridge this, our second pathway calls for \textbf{structural interventions in our academic venues}. We laud IASEAI for acting as an inaugural act of unification, but we advocate that the venue and its organizers lean into interdisciplinary thought---pioneering joint conference tracks, workshops, and doctoral consortia that require co-authorship between technical and social scientists. By creating spaces where collaboration is not just encouraged but expected---and, critically, evaluated by mixed review panels---we can lower the high social and professional cost of crossing the disciplinary divide our data reveals.

\paragraph{3. Integrative Research Methodologies.} Finally, we propose a pathway of \textbf{methodological synthesis}. The tools of one community can solve the problems of the other. The mechanistic interpretability techniques developed in AI safety, for example, are powerful instruments for conducting the deep algorithmic audits called for by AI ethics. Conversely, the participatory design methods from ethics offer a robust framework for operationalizing the "human values" that scalable oversight in AI safety aims to align with. We advocate for research that explicitly couples these approaches as a concrete means of forging a unified, socio-technical practice of alignment.

\bibliographystyle{acm}
\bibliography{cites}

@inproceedings{vijjini-etal-2025-exploring,
    title = "Exploring Safety-Utility Trade-Offs in Personalized Language Models",
    author = "Vijjini, Anvesh Rao  and
      Basu Roy Chowdhury, Somnath  and
      Chaturvedi, Snigdha",
    editor = "Chiruzzo, Luis  and
      Ritter, Alan  and
      Wang, Lu",
    booktitle = "Proceedings of the 2025 Conference of the Nations of the Americas Chapter of the Association for Computational Linguistics: Human Language Technologies (Volume 1: Long Papers)",
    month = apr,
    year = "2025",
    address = "Albuquerque, New Mexico",
    publisher = "Association for Computational Linguistics",
    url = "https://aclanthology.org/2025.naacl-long.565/",
    doi = "10.18653/v1/2025.naacl-long.565",
    pages = "11316--11340",
    ISBN = "979-8-89176-189-6"
}

@misc{li2025smarterllmssaferexploring,
      title={Are Smarter LLMs Safer? Exploring Safety-Reasoning Trade-offs in Prompting and Fine-Tuning}, 
      author={Ang Li and Yichuan Mo and Mingjie Li and Yifei Wang and Yisen Wang},
      year={2025},
      eprint={2502.09673},
      archivePrefix={arXiv},
      primaryClass={cs.CL},
      url={https://arxiv.org/abs/2502.09673}, 
}

@article{shen2024jailbreak,
  title={Jailbreak antidote: Runtime safety-utility balance via sparse representation adjustment in large language models},
  author={Shen, Guobin and Zhao, Dongcheng and Dong, Yiting and He, Xiang and Zeng, Yi},
  journal={arXiv preprint arXiv:2410.02298},
  year={2024}
}

@misc{elias_ai_2025,
	title = {{AI} research takes a backseat to profits as {Silicon} {Valley} prioritizes products over safety, experts say},
	url = {https://www.cnbc.com/2025/05/14/meta-google-openai-artificial-intelligence-safety.html},
	language = {en},
	urldate = {2025-09-28},
	journal = {CNBC},
	author = {Elias, Jonathan Vanian and Jennifer, Hayden Field},
	month = may,
	year = {2025},
	note = {Section: AI Effect},
}

@article{borenstein_ai_2021,
	title = {{AI} {Ethics}: {A} {Long} {History} and a {Recent} {Burst} of {Attention}},
	volume = {54},
	issn = {0018-9162},
	shorttitle = {{AI} {Ethics}},
	url = {https://www.computer.org/csdl/magazine/co/2021/01/09321834/1qmbkXCazy8},
	doi = {10.1109/MC.2020.3034950},
	language = {English},
	number = {01},
	urldate = {2025-09-28},
	journal = {Computer},
	author = {Borenstein, Jason and Grodzinsky, Frances S. and Howard, Ayanna and Miller, Keith W. and Wolf, Marty J.},
	month = jan,
	year = {2021},
	note = {Publisher: IEEE Computer Society},
	pages = {96--102},
}

@misc{amodei_concrete_2016,
	title = {Concrete {Problems} in {AI} {Safety}},
	url = {http://arxiv.org/abs/1606.06565},
	doi = {10.48550/arXiv.1606.06565},
	urldate = {2025-09-28},
	publisher = {arXiv},
	author = {Amodei, Dario and Olah, Chris and Steinhardt, Jacob and Christiano, Paul and Schulman, John and Mané, Dan},
	month = jul,
	year = {2016},
	note = {arXiv:1606.06565 [cs]},
}

@misc{shi_large_2024,
	title = {Large {Language} {Model} {Safety}: {A} {Holistic} {Survey}},
	shorttitle = {Large {Language} {Model} {Safety}},
	url = {http://arxiv.org/abs/2412.17686},
	doi = {10.48550/arXiv.2412.17686},
	urldate = {2025-09-28},
	publisher = {arXiv},
	author = {Shi, Dan and Shen, Tianhao and Huang, Yufei and Li, Zhigen and Leng, Yongqi and Jin, Renren and Liu, Chuang and Wu, Xinwei and Guo, Zishan and Yu, Linhao and Shi, Ling and Jiang, Bojian and Xiong, Deyi},
	month = dec,
	year = {2024},
	note = {arXiv:2412.17686 [cs]},
}

@misc{lin_ai_2025,
	title = {{AI} {Safety} vs. {AI} {Security}: {Demystifying} the {Distinction} and {Boundaries}},
	shorttitle = {{AI} {Safety} vs. {AI} {Security}},
	url = {http://arxiv.org/abs/2506.18932},
	doi = {10.48550/arXiv.2506.18932},
	urldate = {2025-09-28},
	publisher = {arXiv},
	author = {Lin, Zhiqiang and Sun, Huan and Shroff, Ness},
	month = jun,
	year = {2025},
	note = {arXiv:2506.18932 [cs]},
}

@misc{gu_survey_2024,
	title = {A {Survey} on {Responsible} {Generative} {AI}: {What} to {Generate} and {What} {Not}},
	shorttitle = {A {Survey} on {Responsible} {Generative} {AI}},
	url = {http://arxiv.org/abs/2404.05783},
	doi = {10.48550/arXiv.2404.05783},
	urldate = {2025-09-28},
	publisher = {arXiv},
	author = {Gu, Jindong},
	month = sep,
	year = {2024},
	note = {arXiv:2404.05783 [cs]},
}

@misc{swoboda_examining_2025,
	title = {Examining {Popular} {Arguments} {Against} {AI} {Existential} {Risk}: {A} {Philosophical} {Analysis}},
	shorttitle = {Examining {Popular} {Arguments} {Against} {AI} {Existential} {Risk}},
	url = {http://arxiv.org/abs/2501.04064},
	doi = {10.48550/arXiv.2501.04064},
	urldate = {2025-09-28},
	publisher = {arXiv},
	author = {Swoboda, Torben and Uuk, Risto and Lauwaert, Lode and Rebera, Andrew P. and Oimann, Ann-Katrien and Chomanski, Bartlomiej and Prunkl, Carina},
	month = jan,
	year = {2025},
	note = {arXiv:2501.04064 [cs]},
}

@misc{bengio_international_2025,
	title = {International {AI} {Safety} {Report}},
	url = {http://arxiv.org/abs/2501.17805},
	doi = {10.48550/arXiv.2501.17805},
	urldate = {2025-10-04},
	publisher = {arXiv},
	author = {Bengio, Yoshua and Mindermann, Sören and Privitera, Daniel and Besiroglu, Tamay and Bommasani, Rishi and Casper, Stephen and Choi, Yejin and Fox, Philip and Garfinkel, Ben and Goldfarb, Danielle and Heidari, Hoda and Ho, Anson and Kapoor, Sayash and Khalatbari, Leila and Longpre, Shayne and Manning, Sam and Mavroudis, Vasilios and Mazeika, Mantas and Michael, Julian and Newman, Jessica and Ng, Kwan Yee and Okolo, Chinasa T. and Raji, Deborah and Sastry, Girish and Seger, Elizabeth and Skeadas, Theodora and South, Tobin and Strubell, Emma and Tramèr, Florian and Velasco, Lucia and Wheeler, Nicole and Acemoglu, Daron and Adekanmbi, Olubayo and Dalrymple, David and Dietterich, Thomas G. and Felten, Edward W. and Fung, Pascale and Gourinchas, Pierre-Olivier and Heintz, Fredrik and Hinton, Geoffrey and Jennings, Nick and Krause, Andreas and Leavy, Susan and Liang, Percy and Ludermir, Teresa and Marda, Vidushi and Margetts, Helen and McDermid, John and Munga, Jane and Narayanan, Arvind and Nelson, Alondra and Neppel, Clara and Oh, Alice and Ramchurn, Gopal and Russell, Stuart and Schaake, Marietje and Schölkopf, Bernhard and Song, Dawn and Soto, Alvaro and Tiedrich, Lee and Varoquaux, Gaël and Yao, Andrew and Zhang, Ya-Qin and Albalawi, Fahad and Alserkal, Marwan and Ajala, Olubunmi and Avrin, Guillaume and Busch, Christian and Carvalho, André Carlos Ponce de Leon Ferreira de and Fox, Bronwyn and Gill, Amandeep Singh and Hatip, Ahmet Halit and Heikkilä, Juha and Jolly, Gill and Katzir, Ziv and Kitano, Hiroaki and Krüger, Antonio and Johnson, Chris and Khan, Saif M. and Lee, Kyoung Mu and Ligot, Dominic Vincent and Molchanovskyi, Oleksii and Monti, Andrea and Mwamanzi, Nusu and Nemer, Mona and Oliver, Nuria and Portillo, José Ramón López and Ravindran, Balaraman and Rivera, Raquel Pezoa and Riza, Hammam and Rugege, Crystal and Seoighe, Ciarán and Sheehan, Jerry and Sheikh, Haroon and Wong, Denise and Zeng, Yi},
	month = jan,
	year = {2025},
	note = {arXiv:2501.17805 [cs]},
}

@article{judge_when_2025,
	title = {When code isn’t law: rethinking regulation for artificial intelligence},
	volume = {44},
	issn = {1449-4035},
	shorttitle = {When code isn’t law},
	url = {https://doi.org/10.1093/polsoc/puae020},
	doi = {10.1093/polsoc/puae020},
	number = {1},
	urldate = {2025-10-04},
	journal = {Policy and Society},
	author = {Judge, Brian and Nitzberg, Mark and Russell, Stuart},
	month = apr,
	year = {2025},
	pages = {85--97},
}

@misc{noauthor_analyzing_2019,
	title = {Analyzing preprints: {The} challenges of working with metadata from {arXiv}’s {Quantitative} {Biology} section},
	shorttitle = {Analyzing preprints},
	url = {https://www.scholcommlab.ca/2019/11/07/preprints-challenges-part-four/},
	language = {en-CA},
	urldate = {2025-10-10},
	journal = {Scholarly Communications Lab {\textbar} ScholCommLab},
	month = nov,
	year = {2019},
}

@book{bostrom_superintelligence_2014,
	title = {Superintelligence: {Paths}, {Dangers}, {Strategies}},
	isbn = {978-0-19-166682-7},
	url = {https://books.google.com/books?id=C-_8AwAAQBAJ},
	publisher = {OUP Oxford},
	author = {Bostrom, N.},
	year = {2014},
	lccn = {2013955152},
}

@article{behrent2013foucault,
  title={Foucault and technology},
  author={Behrent, Michael C},
  journal={History and Technology},
  volume={29},
  number={1},
  pages={54-104},
  year={2013},
  publisher={Taylor \& Francis}
}

@inproceedings{buolamwini2018gender,
  title={Gender shades: Intersectional accuracy disparities in commercial gender classification},
  author={Buolamwini, Joy and Gebru, Timnit},
  booktitle={Conference on fairness, accountability and transparency},
  pages={77--91},
  year={2018},
  organization={PMLR}
}

@inproceedings{bender2021dangers,
  title={On the dangers of stochastic parrots: Can language models be too big?},
  author={Bender, Emily M and Gebru, Timnit and McMillan-Major, Angelina and Shmitchell, Shmargaret},
  booktitle={Proceedings of the 2021 ACM conference on fairness, accountability, and transparency},
  pages={610--623},
  year={2021}
}

@article{munn2023uselessness,
  title={The uselessness of AI ethics},
  author={Munn, Luke},
  journal={AI and Ethics},
  volume={3},
  number={3},
  pages={869--877},
  year={2023},
  publisher={Springer}
}

@misc{dblp2025dump,
  title        = {DBLP Computer Science Bibliography (2025 XML dump)},
  howpublished = {\url{https://dblp.org/xml/}},
  note         = {Data snapshot used for author/venue metadata, accessed 2025-10-11},
  year         = {2025}
}

@article{horta_homophily_2022,
	title = {Homophily in higher education research: a perspective based on co-authorships},
	volume = {127},
	issn = {1588-2861},
	shorttitle = {Homophily in higher education research},
	url = {https://doi.org/10.1007/s11192-021-04227-z},
	doi = {10.1007/s11192-021-04227-z},
	language = {en},
	number = {1},
	urldate = {2025-10-11},
	journal = {Scientometrics},
	author = {Horta, Hugo and Feng, Shihui and Santos, João M.},
	month = jan,
	year = {2022},
	pages = {523--543},
}

@article{haghani_what_2023,
	title = {What makes an informative and publication-worthy scientometric analysis of literature: {A} guide for authors, reviewers and editors},
	volume = {22},
	issn = {2590-1982},
	shorttitle = {What makes an informative and publication-worthy scientometric analysis of literature},
	url = {https://www.sciencedirect.com/science/article/pii/S2590198223002038},
	doi = {10.1016/j.trip.2023.100956},
	urldate = {2025-10-11},
	journal = {Transportation Research Interdisciplinary Perspectives},
	author = {Haghani, Milad},
	month = nov,
	year = {2023},
	pages = {100956},
}

@article{khanam2023homophily,
  title={The homophily principle in social network analysis: A survey},
  author={Khanam, Kazi Zainab and Srivastava, Gautam and Mago, Vijay},
  journal={Multimedia Tools and Applications},
  volume={82},
  number={6},
  pages={8811--8854},
  year={2023},
  publisher={Springer}
}

@phdthesis{saxena2024shuffling,
  title={The Shuffling Effect: Vertex Label Error’s Impact on Hypothesis Testing, Classification, and Clustering in Graph Data},
  author={Saxena, Ayushi},
  year={2024},
  school={University of Maryland, College Park}
}

@article{vavsa2022null,
  title={Null models in network neuroscience},
  author={V{\'a}{\v{s}}a, Franti{\v{s}}ek and Mi{\v{s}}i{\'c}, Bratislav},
  journal={Nature Reviews Neuroscience},
  volume={23},
  number={8},
  pages={493--504},
  year={2022},
  publisher={Nature Publishing Group UK London}
}

@article{newman2004finding,
  title={Finding and evaluating community structure in networks},
  author={Newman, Mark EJ and Girvan, Michelle},
  journal={Physical review E},
  volume={69},
  number={2},
  pages={026113},
  year={2004},
  publisher={APS}
}

@misc{tahamtan2019citationcountsmeasureupdated,
      title={What Do Citation Counts Measure? An Updated Review of Studies on Citations in Scientific Documents Published between 2006 and 2018}, 
      author={Iman Tahamtan and Lutz Bornmann},
      year={2019},
      eprint={1906.04588},
      archivePrefix={arXiv},
      primaryClass={cs.DL},
      url={https://arxiv.org/abs/1906.04588}, 
}

@article{newman2004coauthorship,
  title={Coauthorship networks and patterns of scientific collaboration},
  author={Newman, Mark EJ},
  journal={Proceedings of the national academy of sciences},
  volume={101},
  number={suppl\_1},
  pages={5200--5205},
  year={2004},
  publisher={National Academy of Sciences}
}

@misc{acl2025dump,
  title        = {ACL Anthology Bibliography (2025 XML dump)},
  howpublished = {\url{https://aclanthology.org/anthology+abstracts.bib.gz
}},
  note         = {ACL snapshot used for author/venue metadata, accessed 2025-10-11},
  year         = {2025}
}

@article{sadek2025challenges,
  title={Challenges of responsible AI in practice: scoping review and recommended actions},
  author={Sadek, Malak and Kallina, Emma and Bohn{\'e}, Thomas and Mougenot, C{\'e}line and Calvo, Rafael A and Cave, Stephen},
  journal={AI \& society},
  volume={40},
  number={1},
  pages={199--215},
  year={2025},
  publisher={Springer}
}

@book{d2023data,
  title={Data feminism},
  author={D'ignazio, Catherine and Klein, Lauren F},
  year={2023},
  publisher={MIT press}
}

@incollection{benjamin2023race,
  title={Race after technology},
  author={Benjamin, Ruha},
  booktitle={Social Theory Re-Wired},
  pages={405--415},
  year={2023},
  publisher={Routledge}
}

@article{gebru_effective_nodate,
	title = {Effective {Altruism} {Is} {Pushing} a {Dangerous} {Brand} of ‘{AI} {Safety}’},
	issn = {1059-1028},
	url = {https://www.wired.com/story/effective-altruism-artificial-intelligence-sam-bankman-fried/},
	language = {en-US},
	urldate = {2025-10-10},
	journal = {Wired},
	author = {Gebru, Timnit},
	note = {Section: tags},
}

@inproceedings{mosca2022shap,
  title={SHAP-based explanation methods: a review for NLP interpretability},
  author={Mosca, Edoardo and Szigeti, Ferenc and Tragianni, Stella and Gallagher, Daniel and Groh, Georg},
  booktitle={Proceedings of the 29th international conference on computational linguistics},
  pages={4593--4603},
  year={2022}
}

@article{novelli2024accountability,
  title={Accountability in artificial intelligence: What it is and how it works},
  author={Novelli, Claudio and Taddeo, Mariarosaria and Floridi, Luciano},
  journal={Ai \& Society},
  volume={39},
  number={4},
  pages={1871--1882},
  year={2024},
  publisher={Springer}
}

@misc{mehrabi2022surveybiasfairnessmachine,
      title={A Survey on Bias and Fairness in Machine Learning}, 
      author={Ninareh Mehrabi and Fred Morstatter and Nripsuta Saxena and Kristina Lerman and Aram Galstyan},
      year={2022},
      eprint={1908.09635},
      archivePrefix={arXiv},
      primaryClass={cs.LG},
      url={https://arxiv.org/abs/1908.09635}, 
}

@article{nath2023posthumanism,
  title={From posthumanism to ethics of artificial intelligence},
  author={Nath, Rajakishore and Manna, Riya},
  journal={AI \& SOCIETY},
  volume={38},
  number={1},
  pages={185--196},
  year={2023},
  publisher={Springer}
}

@misc{ren2024safetywashingaisafetybenchmarks,
      title={Safetywashing: Do AI Safety Benchmarks Actually Measure Safety Progress?}, 
      author={Richard Ren and Steven Basart and Adam Khoja and Alice Gatti and Long Phan and Xuwang Yin and Mantas Mazeika and Alexander Pan and Gabriel Mukobi and Ryan H. Kim and Stephen Fitz and Dan Hendrycks},
      year={2024},
      eprint={2407.21792},
      archivePrefix={arXiv},
      primaryClass={cs.LG},
      url={https://arxiv.org/abs/2407.21792}, 
}

@book{yudkowsky_if_2025,
	title = {If {Anyone} {Builds} {It}, {Everyone} {Dies}: {Why} {Superhuman} {AI} {Would} {Kill} {Us} {All}},
	isbn = {978-0-316-59566-7},
	url = {https://books.google.com/books?id=8ZNLEQAAQBAJ},
	publisher = {Little, Brown},
	author = {Yudkowsky, E. and Soares, N.},
	year = {2025},
}

@misc{bereska2024mechanisticinterpretabilityaisafety,
      title={Mechanistic Interpretability for AI Safety -- A Review}, 
      author={Leonard Bereska and Efstratios Gavves},
      year={2024},
      eprint={2404.14082},
      archivePrefix={arXiv},
      primaryClass={cs.AI},
      url={https://arxiv.org/abs/2404.14082}, 
}

@misc{greenblatt2024aicontrolimprovingsafety,
      title={AI Control: Improving Safety Despite Intentional Subversion}, 
      author={Ryan Greenblatt and Buck Shlegeris and Kshitij Sachan and Fabien Roger},
      year={2024},
      eprint={2312.06942},
      archivePrefix={arXiv},
      primaryClass={cs.LG},
      url={https://arxiv.org/abs/2312.06942}, 
}

@article{key11,
  author = {Pearce, Hammond and others},
  title = {Asleep at the Keyboard? Assessing the Security of GitHub Copilot's Code Contributions},
  year = {2022},
  journal = {arXiv preprint arXiv:2108.09293}
}

@misc{key15,
  author = {Bengio, Yoshua},
  title = {A plan to keep AI safe},
  year = {2025},
  howpublished = {TED Talk},
  url = {https://www.youtube.com/watch?v=qe9QSCF-d88}
}

@misc{key16,
  author = {Bostrom, Nick},
  title = {What happens when our computers get smarter than we are?},
  year = {2015},
  howpublished = {TED Talk},
  url = {https://www.ted.com/talks/nick_bostrom_what_happens_when_our_computers_get_smarter_than_we_are}
}

@misc{key19,
  author = {Luccioni, Sasha},
  title = {AI is dangerous, but not for the reasons you think},
  year = {2023},
  howpublished = {TED Talk},
  url = {https://www.ted.com/talks/sasha_luccioni_ai_is_dangerous_but_not_for_the_reasons_you_think}
}

@article{key21,
  author = {Memarian, Bahar and Doleck, Tenzin},
  title = {Fairness, Accountability, Transparency, and Ethics (FATE) in AI},
  year = {2023},
  journal = {Computers and Education: Artificial Intelligence},
  volume = {5},
  pages = {100163}
}

@inproceedings{key24,
  author = {Deck, Luca and others},
  title = {A Critical Survey on Fairness Benefits of Explainable AI},
  booktitle = {Proceedings of the 2024 ACM Conference on Fairness, Accountability, and Transparency (FAccT '24)},
  year = {2024},
  pages = {105--117}
}

@misc{key25,
  author = {{Anthropic}},
  title = {HH-RLHF Repository},
  year = {n.d.},
  howpublished = {\url{https://github.com/anthropics/hh-rlhf}},
  note = {Accessed: 2025-10-10}
}

@inproceedings{key27,
  author = {Verma, Sahil and Rubin, Julia},
  title = {Fairness definitions explained},
  booktitle = {Proceedings of the International Workshop on Software Fairness},
  year = {2018},
  pages = {1--7}
}

@article{key28,
  author = {{Frontiers in Human Dynamics}},
  title = {Transparency and accountability in AI systems: A review of legal and ethical challenges},
  year = {2024},
  journal = {Frontiers in Human Dynamics},
  volume = {6},
  howpublished = {\url{https://www.frontiersin.org/journals/human-dynamics/articles/10.3389/fhumd.2024.1421273/full}}
}

@techreport{key32,
  author = {{UNESCO}},
  title = {Recommendation on the Ethics of Artificial Intelligence},
  year = {2021},
  institution = {UNESCO},
  howpublished = {\url{https://www.unesco.org/en/artificial-intelligence/recommendation-ethics}}
}

@misc{key41,
  author = {{Google Research}},
  title = {VaultGemma: The world's most capable differentially private LLM},
  year = {2024},
  howpublished = {\url{https://research.google/blog/vaultgemma-the-worlds-most-capable-differentially-private-llm/}},
  note = {Accessed: 2025-10-10}
}

@misc{key50,
  author = {Alabi, Moses and Holmes, Tobi},
  title = {AI Safety and Ethics: Developing Robust Frameworks for Ethical AI Development and Deployment},
  year = {2024},
  howpublished = {ResearchGate},
  url = {https://www.researchgate.net/publication/383283895}
}

@misc{key52,
  author = {{The AI Safety Book}},
  title = {AI Safety, Ethics and Society},
  year = {2024},
  howpublished = {\url{https://www.aisafetybook.com/}},
  note = {Accessed: 2025-10-10}
}

@misc{key62,
  author = {{Brookings Institution}},
  title = {With AI, we need both competition and safety},
  year = {2024},
  howpublished = {\url{https://www.brookings.edu/articles/with-ai-we-need-both-competition-and-safety/}},
  note = {Accessed: 2025-10-10}
}

@misc{key66,
  author = {{McKinsey \& Company}},
  title = {As gen AI advances, regulators and risk functions rush to keep pace},
  year = {2025},
  howpublished = {\url{https://www.mckinsey.com/capabilities/risk-and-resilience/our-insights/as-gen-ai-advances-regulators-and-risk-functions-rush-to-keep-pace}},
  note = {Accessed: 2025-10-10}
}

@misc{key73,
  author = {Durmus, Murat},
  title = {The Difference Between AI Safety, AI Ethics, and Responsible AI},
  year = {2024},
  howpublished = {Medium},
  url = {https://murat-durmus.medium.com/the-difference-between-ai-safety-ai-ethics-and-responsible-ai-8296306af427}
}

@misc{key98,
  author = {{Alignment Forum}},
  title = {Outer vs. Inner Misalignment: Three Framings},
  year = {2022},
  howpublished = {\url{https://www.alignmentforum.org/posts/poyshiMEhJsAuifKt/outer-vs-inner-misalignment-three-framings-1}},
  note = {Accessed: 2025-10-10}
}

@article{key102,
  author = {Buolamwini, Joy},
  title = {Unmasking AI: My Mission to Protect What Is Human in a World of Machines},
  year = {2023},
  journal = {MIT Sloan Management Review}
}

\appendix

\section{Dataset Preprocessing}
\label{filter_sec}

\subsection{Abstract Enrichment Coverage}

\begin{table}[htbp!]
\centering
\caption{Abstract Enrichment Coverage by Conference}
\label{tab:enrichment_coverage_yearly}
\begin{tabular}{@{}lrrr@{}}
\toprule
Conference & Total Papers & Papers w/ Abstracts & Final Coverage \\
\midrule
AAAI & 14,137 & 13,883 & 98.2\% \\
AIES & 583 & 578 & 99.1\% \\
FAT & 890 & 885 & 99.4\% \\
ICLR & 10,599 & 10,546 & 99.5\% \\
ICML & 7,979 & 7,979 & 100.0\% \\
NEURIPS & 15,407 & 14,606 & 94.8\% \\
SATML & 130 & 127 & 97.7\% \\
\midrule
\textbf{Total} & \textbf{49,725} & \textbf{48,604} & \textbf{97.7\%} \\
\bottomrule
\end{tabular}
\end{table}

\subsection{Step One: Filtering by Keywords}
\subsubsection{Creating Keyword Sets}
The initial keyword set was generated by manually analyzing the terminology used in foundational surveys and texts in each field. For Safety, we consult recent surveys which consider the five anchoring problems in \cite{amodei_concrete_2016}, such as \cite{key52,shi_large_2024,key15}. For Ethics research, we consider surveys on Responsible Generative AI \cite{gu_survey_2024}, the tradition of ``FATE'' \cite{mehrabi2022surveybiasfairnessmachine,key21,sadek2025challenges}, and \cite{benjamin2023race,judge_when_2025} as references for regulatory and governmental literature. 

Furthermore, we devised a hierarchical strategy that would span reasonable categories of literature expected in collected venues. We used header markers from surveys, as well as language in introductory content, to capture technical, theoretical, and applied domains that would like to relevant research. To ensure face validity and mitigate author bias, the keyword lists and their thematic categorizations were independently reviewed by two graduate researchers with expertise in AI safety and AI ethics, respectively. Their feedback was incorporated into the final version, which can be found in \ref{tab:keyword_taxonomy}.

\subsubsection{Keywords}
\begin{table*}[h!]
\centering
\caption{Keyword Taxonomy for Classifying AI Safety and AI Ethics Research}
\label{tab:keyword_taxonomy}
\begin{tabularx}{\textwidth}{@{} X X @{}}
\toprule
\textbf{AI Safety Keywords} & \textbf{AI Ethics Keywords} \\
\midrule
\textbf{Agentic Risk \& Loss of Control} & \textbf{Core Ethics, Accountability \& Governance} \\
AI control, loss of control, agentic AI, autonomous systems, corrigibility, unintended objectives, unforeseen behavior, AI containment, shutdown problem, oracle AI, tool AI, self-replication. & AI ethics, machine ethics, algorithmic accountability, AI governance, AI regulation, ethical AI, responsible AI, trustworthy AI, human-in-the-loop, meaningful human control, sociotechnical perspective. \\
\addlinespace
\textbf{Goal Misalignment \& Instrumental Goals} & \textbf{Bias, Fairness \& Equity} \\
Goal misalignment, value misalignment, objective misspecification, reward hacking, specification gaming, Goodhart's law, instrumental goals, power-seeking, self-preservation, resource acquisition. & Algorithmic bias, data bias, social bias, systemic bias, AI fairness, algorithmic fairness, fairness metrics, group fairness, individual fairness, intersectional fairness, AI equity, procedural justice, distributive justice. \\
\addlinespace
\textbf{Emergence, Predictability \& Robustness} & \textbf{Discrimination, Identity \& Societal Harms} \\
Emergent capabilities, emergent behavior, unpredictable AI, black-box models, complex systems failure, phase transitions, discontinuous progress, distributional shift, out-of-distribution, AI robustness. & Algorithmic discrimination, disparate impact, disparate treatment, representational harms, stereotyping, marginalization, vulnerable populations, protected class, racial bias, gender bias, racism, sexism. \\
\addlinespace
\textbf{Deception \& Evasion of Oversight} & \textbf{Privacy, Surveillance \& Power Dynamics} \\
AI deception, strategic manipulation, sandbagging, unfaithful reasoning, truthfulness, honesty, jailbreaking, prompt hacking, prompt injection, obfuscated instructions. & Data privacy, algorithmic privacy, surveillance technology, facial recognition, biometric surveillance, social scoring, predictive policing, data exploitation, power dynamics, power asymmetry, digital colonialism. \\
\addlinespace
\textbf{Safety Evaluations \& Red Teaming} & \textbf{Human Agency, Labor \& Well-being} \\
Safety evaluation, dangerous capabilities evaluation, red teaming, red-team, adversarial testing, honey pots, model organism, behavioral evaluations, automated evaluations, testing, validation. & Human dignity, human autonomy, human rights, informed consent, cognitive automation, attention economy, job automation, economic displacement, labor rights, worker surveillance. \\
\addlinespace
\textbf{Catastrophic \& Existential Outcomes} & \textbf{Information Integrity \& Content} \\
Existential risk, global catastrophic risk, AI-driven catastrophe, AI weaponization, AI misuse, dual-use AI, AI arms race, multipolar dynamics, race to the bottom. & Misinformation, disinformation, deepfakes, synthetic content, content moderation, hate speech, online harassment, political polarization, intellectual property, copyright. \\
\addlinespace
\textbf{Control Strategies \& Safeguards} & \textbf{Mitigation \& Responsible Design} \\
AI safety, AI alignment research, guardrails, safety filters, input/output filtering, process supervision, outcome supervision, scalable oversight, human feedback, capability control, fail-safe. & Bias detection, bias mitigation, debiasing, fairness-aware machine learning, value-sensitive design, participatory AI, algorithmic impact assessments, algorithmic audit, transparency, data stewardship, environmental justice. \\
\addlinespace
\textbf{Interpretability \& Monitoring} & \textbf{Application Domains} \\
AI monitoring, interpretability, explainability, AI transparency, mechanistic interpretability, AI auditing for safety, formal verification, mathematical guarantees. & AI in healthcare, AI in criminal justice, AI in employment, AI in education. \\
\bottomrule
\end{tabularx}
\end{table*}

\subsubsection{Validating Keyword Search}

To empirically validate the utility of keyword-based filtering, we conduct a test using 225 hand-labeled examples. This set consists of 75 gold-standard ethics, 75 gold-standard safety, and 75 unrelated papers. Gold-standard entries were randomly sampled from the surveys used to collect keywords, provided that the entry was published during or after 2020. The unrelated papers were sampled at large from our NeurIPS dataset.

\begin{figure}[h]
    \includegraphics[width=0.5\linewidth]{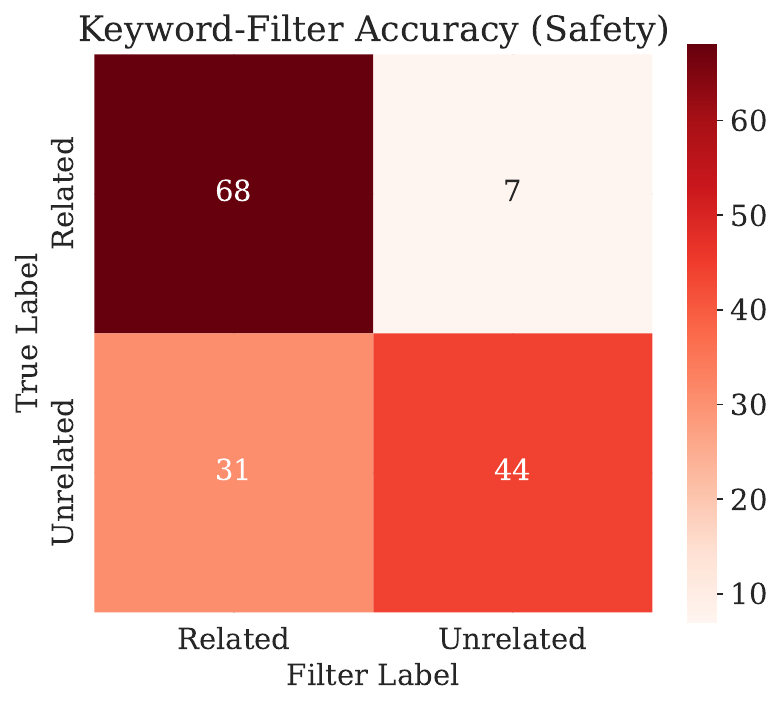}
    \hfill
    \includegraphics[width=0.5\linewidth]{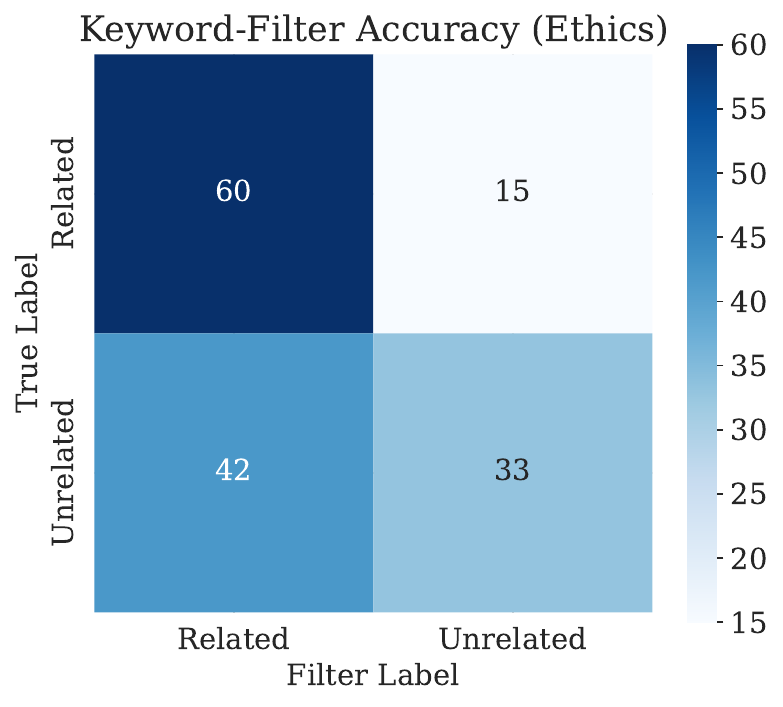}

    \caption{Confusion Matrices of Keyword-Based Classifiers show a high TPR.} \    
    \label{fig:placeholder}
\end{figure}

We achieve a high TPR of 90.67\% on safety papers, and 80\% on ethics papers \ref{fig:placeholder}. However, the False Positive Rate was quite high---41.3\% on safety and 56\% on ethics. This motivated the introduction of a second filter. The ethics papers in particular pose a challenge as researchers appear to adopt a wider variety of meanings that may not reflect the scoping definitions of surveys.

\subsection{Step Two: LM-Based Filtering}

\subsubsection{Specification}

Noticing the relatively high false-positive rates in our validation, we utilize a language model to filter out papers that may not be relevant. Specifically, we dispatch batches of eight papers, all from the same conference, with a yes/no, confidence, and reasoning request in output (see \ref{llm-filter-prompt}. We used Gemini-2.5-Flash, with chunks of eight processed in parallel.

\begin{figure}[h!]
\begin{tcolorbox}[
  colback=black!5!white,  
  colframe=black!75!white, 
  title=\textbf{LLM Prompt: Academic Paper Filtering},
  fonttitle=\bfseries,
  breakable, 
  pad at break=2mm,
  ]

You are an expert research assistant tasked with filtering academic papers based on their relevance to AI Ethics and AI Safety. Your goal is to determine if a paper's primary contribution falls into one of these two fields.

You will be given the paper's title and abstract. You must return a JSON object with the following structure:
\begin{verbatim}
{
  "include": boolean,
  "category": "ai_ethics" | "ai_safety" | "none",
  "confidence": float (0.0 to 1.0),
  "reasoning": "A brief explanation for your decision."
}
\end{verbatim}

\textbf{Core Task:}
\begin{itemize}
    \item Read the title and abstract carefully.
    \item Decide if the paper's MAIN TOPIC is AI Ethics or AI Safety.
    \item Do not just look for keywords. The CORE FOCUS of the paper is what matters.
\end{itemize}

\textbf{Definitions:}
\begin{enumerate}
    \item \textbf{AI Safety:} Focuses on the technical challenges of ensuring advanced AI systems are robust, reliable, and behave as intended.
    \begin{itemize}
        \item \textbf{Includes:} Value alignment, robustness, interpretability, scalability, corrigibility, and avoiding catastrophic risks or unforeseen negative consequences from highly capable AI systems.
        \item \textbf{Example of a PASS:} A paper on a new method to make large language models less likely to generate harmful content.
        \item \textbf{Example of a FAIL:} A paper on making a system "safe" from traditional cybersecurity threats is NOT AI Safety.
    \end{itemize}
    \vspace{1em} 
    \item \textbf{AI Ethics:} Focuses on the societal, moral, and philosophical implications of AI.
    \begin{itemize}
        \item \textbf{Includes:} Bias and fairness, accountability, transparency, privacy, societal impact, AI governance, and the moral status of AI.
        \item \textbf{Example of a PASS:} A paper analyzing how hiring algorithms can perpetuate gender bias.
        \item \textbf{Example of a FAIL:} A paper that applies AI to solve an ethical problem in another field is NOT an AI Ethics paper.
    \end{itemize}
\end{enumerate}

\textbf{Filtering Rules (IMPORTANT):}
\begin{itemize}
    \item \textbf{INCLUDE (pass):} The paper's primary research contribution is in AI Ethics or AI Safety.
    \item \textbf{EXCLUDE (fail):}
    \begin{itemize}
        \item The paper is merely an \textit{application} of AI to another domain.
        \item The paper is about a dataset or survey without a novel contribution in ethics or safety.
        \item The paper discusses traditional software engineering topics not specific to AI.
        \item The paper is only tangentially related.
    \end{itemize}
\end{itemize}

\textbf{Your Response:}
\begin{itemize}
    \item Your response MUST be a single, valid JSON object.
    \item \texttt{include}: \texttt{true} if the paper should be included, \texttt{false} otherwise.
    \item \texttt{category}: If \texttt{include} is \texttt{true}, specify \texttt{"ai\_ethics"} or \texttt{"ai\_safety"}. If \texttt{false}, use \texttt{"none"}.
    \item \texttt{confidence}: How sure are you of your decision? 1.0 for very sure, 0.5 for uncertain.
    \item \texttt{reasoning}: A concise, one-sentence justification for your decision.
\end{itemize}
\end{tcolorbox}
\caption{Prompt used for LLM-filter}
\label{llm-filter-prompt}
\end{figure}

\subsubsection{Validation}
To validate the LM-based filter, we sample 40 filtering decisions from ACL 2025, NeurIPS 2024, and ICML 2022 each, for a total of 120 samples, and manually annotate our own decision and confidence intervals. Note that any LLM-passed filter queries had already passed keyword search. With two annotators, we achieve a Cohen's kappa of .925 (111 examples). The agreement relative to the language model was .91 (109) and .94 (113 examples). The high rates of agreement motivate our use of the second filter, as it effectively removed false positives from our high-volume conference samples. 

We also run the LLM filter on the full set of manually-selected keyword filters. The LLM accurately classified all but one example, which amounted to a different perspective on prompt definition.

\subsection{Results from Filtering}
\label{filter-app}
\begin{table}[h]
  \caption{Filtering Process Summary}
  \label{filtering-table}
  \centering
  \begin{tabular}{l r r r}
    \toprule
    Conference & Pre-filtering & Post-keyword-filtering & Post-LLM-filtering \\
    \midrule
    AAAI & 14,137 & 1,022 & 779 \\
    AIES & 583 & 560 & 535 \\
    FAT & 890 & 843 & 814 \\
    ACL Anthology & 52,604 & 6,174 & 2,156 \\
    ICLR & 10,599 & 613 & 602 \\
    ICML & 7,979 & 449 & 406 \\
    NeurIPS & 15,407 & 1,970 & 1,038 \\
    SaTML & 130 & 115 & 112 \\
    \midrule
    \textbf{Total} & \textbf{102,329} & \textbf{11,831} & \textbf{6,442} \\
    \bottomrule
  \end{tabular}
\end{table}

\section{Additional Visualizations}
\subsection{Author-based Network}
\begin{figure}[ht!]
    \centering
    \includegraphics[width=0.9\linewidth]{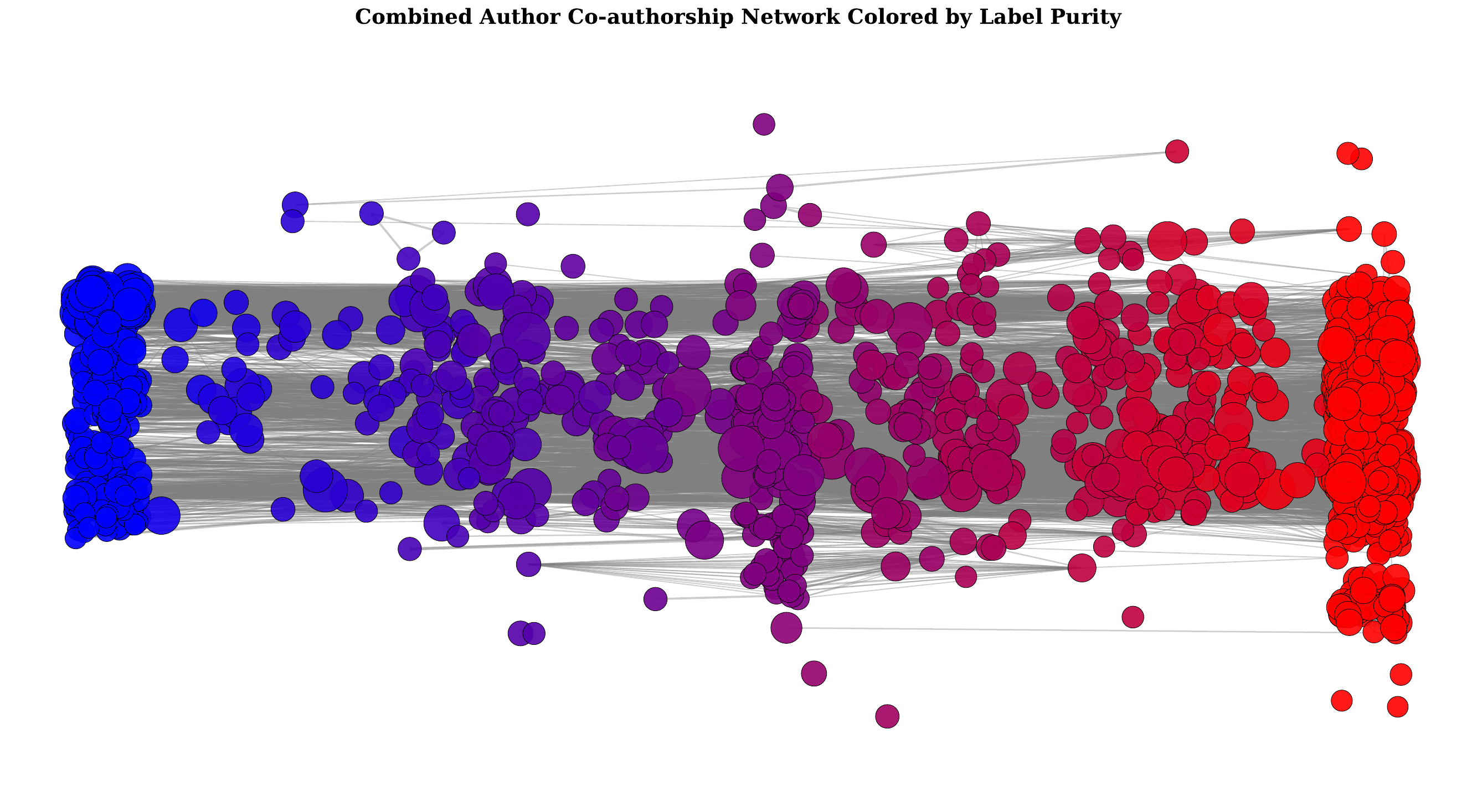}
    \caption{Author-based network similar to Figure \ref{intro}, demonstrating the wider breadth of mixed authors}
    \label{author-network}
\end{figure}
\subsection{Semantic Similarity}
\begin{figure}
\centering
\includegraphics[width=0.9\linewidth]{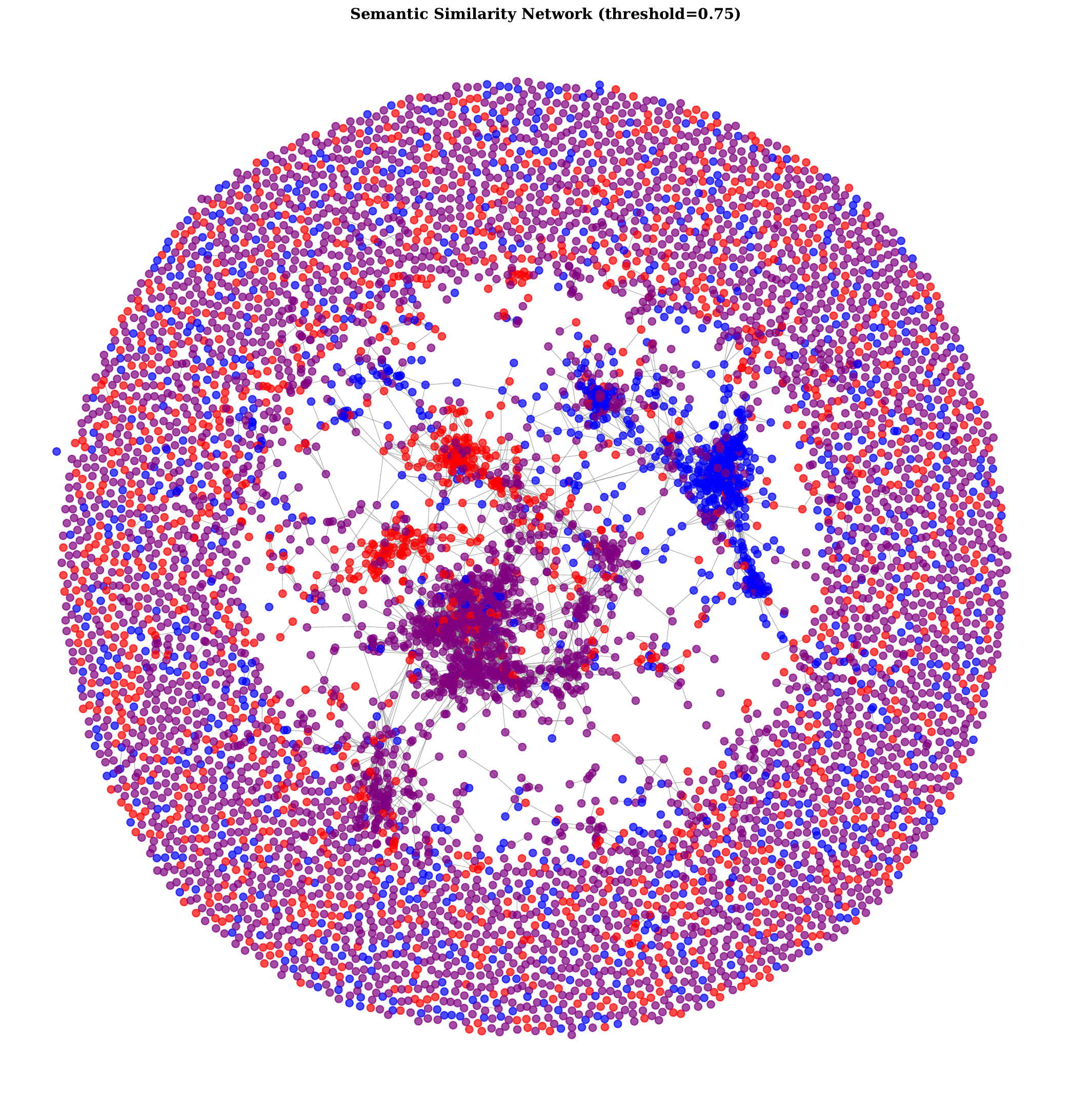}
\caption{Semantic Similarity Network, clustered with k-means at k=10, made with \texttt{all-MiniLM-L6-v2}}
\end{figure}

\section{Experimental Details}
\subsection{Homophily}
\subsection{Bridge Connectivity Tests}
\subsection{Weighted Average Shortest Path Distributions}
We conduct our test by calculated weighted average shortest weighted path for each potential pair---safety to safety, ethics to ethics, and safety-to-safety---without perturbations. The diagram demonstrates the author network, while the values below demonstrate the probability over a uniform selection of author nodes to reach a certain path in a set amount of ``hops''.

\begin{table}[ht!]
\caption{Author reachability within k hops}
\label{tab:wasp_reachability}
\centering
\begin{tabular}{@{}l ccc c c@{}}
\toprule
& \multicolumn{4}{c}{Observed Reachability} & \\
\cmidrule(r){2-5}
Hops (k) & Safety-Safety & Ethics-Ethics & Avg. Within-Group & Safety-Ethics & Separation Delta \\
\midrule
3 & 4.1\% & 2.3\% & 3.2\% & 1.8\% & -1.4\% {\scriptsize(p<0.05)} \\
5 & 23.6\% & 15.0\% & 19.3\% & 16.9\% & -2.4\% {\scriptsize(p<0.01)} \\
7 & 40.0\% & 29.5\% & 34.8\% & 33.8\% & -1.0\% {\scriptsize(p>0.05)} \\
\bottomrule
\end{tabular}
\end{table}

\end{document}